\documentclass[square,numbers]{article}

\usepackage[final]{nips_2017}

\usepackage[utf8]{inputenc} % allow utf-8 input
\usepackage[T1]{fontenc}    % use 8-bit T1 fonts
\usepackage{hyperref}       % hyperlinks
\usepackage{url}            % simple URL typesetting
\usepackage{booktabs}       % professional-quality tables
\usepackage{amsfonts}       % blackboard math symbols
\usepackage{nicefrac}       % compact symbols for 1/2, etc.
\usepackage{microtype}      % microtypography
\usepackage{amsmath}
\usepackage{multicol}
\usepackage{multirow}
\usepackage{hyperref}
\usepackage{url}
\usepackage{graphicx}
\usepackage{caption}
\usepackage{subcaption}
\usepackage{xspace}
\usepackage{enumitem}
\usepackage{dsfont}
\usepackage[usenames,dvipsnames,svgnames,table]{xcolor}

\newcommand{\ds}{Multimodal Attribute Extraction\xspace}
\newcommand{\dsabbrev}{MAE\xspace}

\title{Multimodal Attribute Extraction}

\author{
    Robert L.~Logan IV\\
    University of California \\
    Irvine, CA \\
    \texttt{rlogan@uci.edu}
    \And
    Samuel Humeau \\
    Diffbot \\
    Mountain View, CA \\
    \texttt{sam@diffbot.com}
    \And
    Sameer Singh \\
    University of California\\
    Irvine, CA \\
    \texttt{sameer@uci.edu}
}

\begin{document}

\maketitle

\section{Introduction}

%\textit{WARNING: All results presented in this paper are preliminary. No identification with definitive results is intended or should be inferred. No animals were harmed in the writing of this paper.}

Given the large collections of unstructured and semi-structured data available on the web, there is a crucial need to enable quick and efficient access to the knowledge content within them. 
Traditionally, the field of information extraction has focused on extracting such knowledge from unstructured \emph{text} documents, such as job postings, scientific papers, news articles, and emails.
However, the content on the web increasingly contains more varied types of data, including semi-structured web pages, tables that do not adhere to any schema, photographs, videos, and audio.
Given a query by a user, the appropriate information may appear in any of these different modes, and thus there's a crucial need for methods to construct knowledge bases from different types of data, and more importantly, combine the evidence in order to extract the correct answer.

Motivated by this goal, we introduce the task of \textit{multimodal attribute extraction}.
Provided contextual information about an entity, in the form of any of the modes described above, along with an attribute query, the goal is to extract the corresponding value for that attribute.
While attribute extraction on the domain of text has been well-studied \citep{bing2012unsupervised, Ghani:2006, more2016attribute, putthividhya2011bootstrapped, shinzato2013unsupervised}, to our knowledge this is the first time attribute extraction using a combination of multiple modes of data has been considered.
This introduces additional challenges to the problem, since a multimodal attribute extractor needs to be able to return values provided any kind of evidence, whereas modern attribute extractors treat attribute extraction as a tagging problem and thus only work when attributes occur as a substring of text.

\begin{figure}[tb]
    \includegraphics[width=\textwidth]{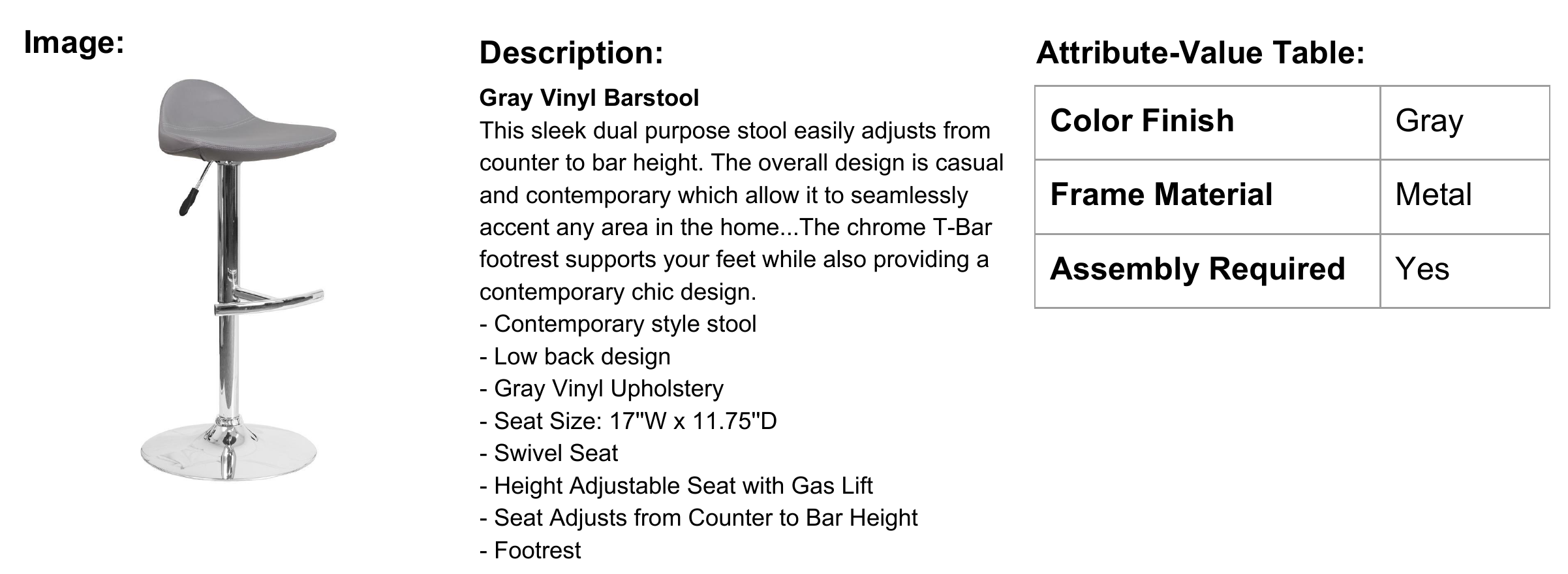}
    \caption{An example item with its descriptions: image, tabular attributes, and textual.}
    \label{fig:example}
\end{figure}

In order to support research on this task, we release the \ds (\dsabbrev) dataset\footnote{The dataset is freely available at: \url{https://rloganiv.github.io/mae}.}, a large dataset containing mixed-media data for over 2.2 million commercial product items, collected from a large number of e-commerce sites using the Diffbot Product API.\footnote{\url{https://www.diffbot.com/products/automatic/product/}}
The collection of items is diverse and includes categories such as electronic products, jewelry, clothing, vehicles, and real estate.
For each item, we provide a textual product description, collection of images, and open-schema table of attribute-value pairs (see Figure~\ref{fig:example} for an example).
The provided attribute-value pairs only provide a very weak source of supervision; where the value might appear in the context is not known, and further, it is not even guaranteed that the value can be extracted from the provided evidence. % isthat can be used to train an attribute extractor in a weakly supervised manner.
In all, there are over 4 million images and 7.6 million attribute-value pairs. 
By releasing such a large dataset, we hope to drive progress on this task similar to how the Penn Treebank~\citep{marcus1993building}, SQuAD~\citep{rajpurkar2016squad}, and Imagenet~\citep{deng2009imagenet} have driven progress on syntactic parsing, question answering, and object recognition, respectively.

To asses the difficulty of the task and the dataset, we first conduct a human evaluation study using Mechanical Turk that demonstrates that all available modes of information are useful for detecting values. 
We also train and provide results for a variety of machine learning models on the dataset. 
We observe that a simple \emph{most-common value classifier}, which always predicts the most-common value for a given attribute, provides a very difficult baseline for more complicated models to beat (33\% accuracy). 
In our current experiments, we are unable to train an image-only classifier that can outperform this simple model, despite using modern neural architectures such as VGG-16\cite{simonyan2014very} and Google's Inception-v3\cite{szegedy2016rethinking}. 
However, we are able to obtain significantly better performance using a text-only classifier (59\% accuracy).
We hope to improve and obtain more accurate models in further research.

\section{Multimodal Product Attribute Extraction}

Since a multimodal attribute extractor needs to be able to return values for attributes which occur in images as well as text, we cannot treat the problem as a labeling problem as is done in the existing approaches to attribute extraction.
We instead define the problem as following:
Given a product $i$ and a query attribute $a$, we need to extract a corresponding value $v$ from the evidence provided for $i$, namely, a textual description of it ($D_i$) and a collection of images ($I_i$).
For example, in Figure~\ref{fig:example}, we observe the image and the description of a product, and examples of some attributes and values of interest.
For training, for a set of product items $\mathcal{I}$, we are given, for each item $i \in \mathcal{I}$, its textual description $D_i$ and the images $I_i$, and a set $A_i$ comprised of attribute-value pairs (i.e. $A_i=\{\langle a_i^j, v_i^j \rangle\}_j$).
%Given a query consisting of an item $i'$ ($i'\notin\mathcal{I}$) and an attribute $a$, the task is to extract the value $v$ such that $\langle a, v \rangle \in A_{i'}$.
In general, the products at query time will not be in $\mathcal{I}$, and we do not assume any fixed ontology for products, attributes, or values.
We evaluate the performance on this task as the accuracy of the predicted value with the observed value, however since there may be multiple \emph{correct} values, we also include hits@$k$ evaluation.

%The benefit of this formulation is that performance on this task can be measured in a well-defined way using accuracy.

\begin{figure}[tb]
\begin{minipage}{.45\textwidth}
    \centering
    \captionof{table}{\dsabbrev dataset statistics.}
    \label{tbl:stats}
    \begin{tabular}{ll} \\
        \toprule
        \# products & 2.2 m \\
        \# images & 4.0 m \\
        \# attribute-value pairs & 7.6 m \\
        \# unique attributes & 2.1 k \\
        \# unique values & 23.6 k \\
        \bottomrule
    \end{tabular}%
\end{minipage}
\begin{minipage}{.45\textwidth}
    \centering
    \includegraphics[width=\textwidth]{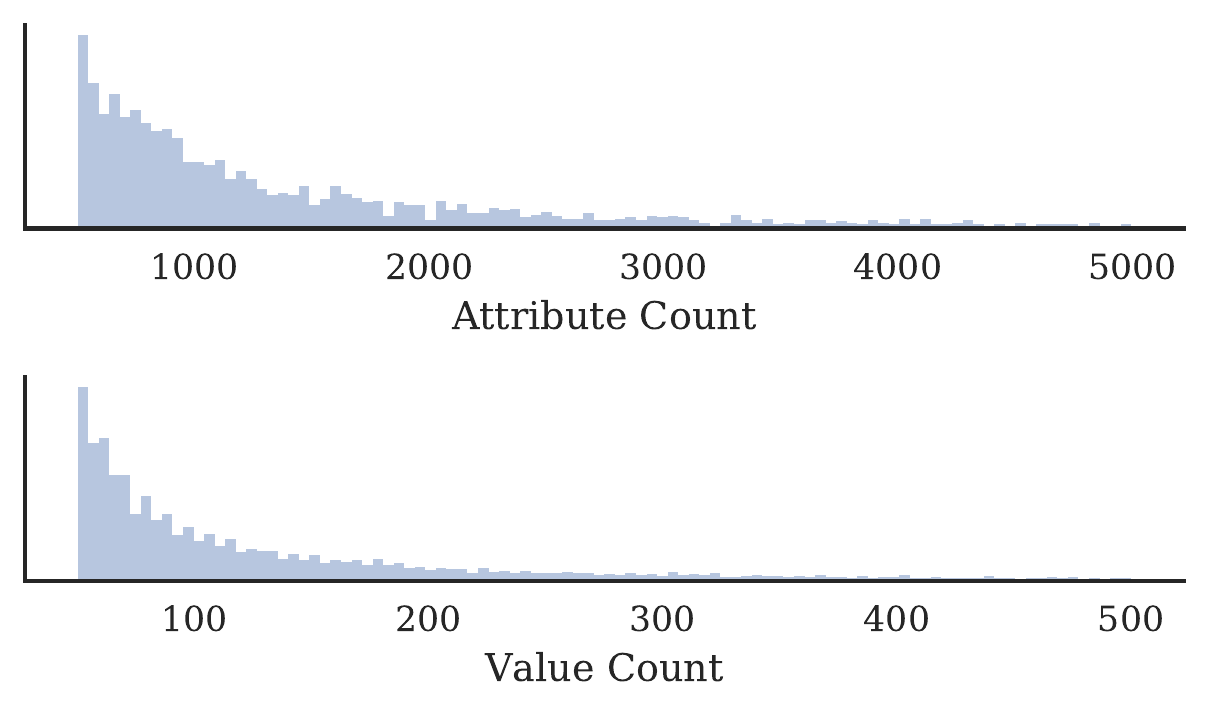}
    \captionof{figure}{Histograms of attribute and value counts.}
    \label{fig:hist}
\end{minipage}\hfill
\end{figure}

\paragraph{The \dsabbrev Dataset}\label{section:mae}
The \dsabbrev dataset is composed of mixed media data for 2.2 million product items, obtained by running the Diffbot Product API on over 20 million web pages from 1068 different commercial websites.
As in the task definition, there is a textual description, set of product images, and open-schema table of product attributes for every item.
The Diffbot API obtains this information using a machine learning based extractor which uses visual, textual and layout features of the fully rendered product webpage.
For example, attribute-value pairs are automatically extracted from tables present on product webpages.
Due to the automated nature of this collection process, there is some noise present in the dataset.
For instance, the same attribute may be represented many different ways (e.g. \textit{Length}, \textit{length}, \textit{len.}).
We use regular-expression based preprocessing to normalize the most common attributes, however, we leave values unnormalized. % in order to preserve overlap with the product descriptions. 
We also remove any attribute-value pairs that satisfy any of the following frequency conditions: the attribute occurs less than 500 times, the value occurs less than 50 times, or the attribute's most common value makes up more than 80\% of the attribute-value pairs. 
% heuristics to filter out the remaining noisy attributes and values:
% \begin{itemize}[nolistsep]
%     \item Any attributes which occur less than 500 times are removed from the dataset.
%     \item Any values which occur less than 50 times are removed from the dataset.
%     \item Any attribute whose most common value makes up more than 80\% of the attribute-value pairs is removed.
% \end{itemize}
The data is split into a training, validation, and test set using an 80-10-10 split.

\paragraph{Mechanical Turk Evaluation}
%A drawback of automatically collecting the dataset using a web scraper is that 
Since the attributes and values have been extracted as they appear on the web sites,
there is no guarantee that the attribute-value pairs appear in either the product images or textual descriptions.
We perform a study using Amazon Mechanical Turk to determine the extent to which this issue affects the dataset, as well as collect a \emph{gold} evaluation dataset of attribute-value pairs that are guaranteed to show up in the context information. % that we can use for evaluation.
Mechanical Turk workers are presented a product's images and textual description, and asked to determine whether they can predict the value for a given product attribute (from a list of choices) using the provided information, and if so, using which pieces of information.
%Positive answers are validated by having workers select the correct value for the attribute query from a brief list.
We use a majority vote to eliminate noise in these annotations. %verify negative answers.
The (preliminary) results of this study suggest that only 42\% of the attribute-value pairs can be found using contextual information.
% Because of the database hack I cannot break out how many are only found in images and only found in text.
Of those, 35\% could be found using the product's image and 70\% could be found using the textual description.
This suggests that while textual descriptions are the most useful mode for attribute extraction, there is still beneficial information contained in images.

\begin{figure} \label{fig:model-diagram}
    \centering
    \includegraphics[width=\textwidth]{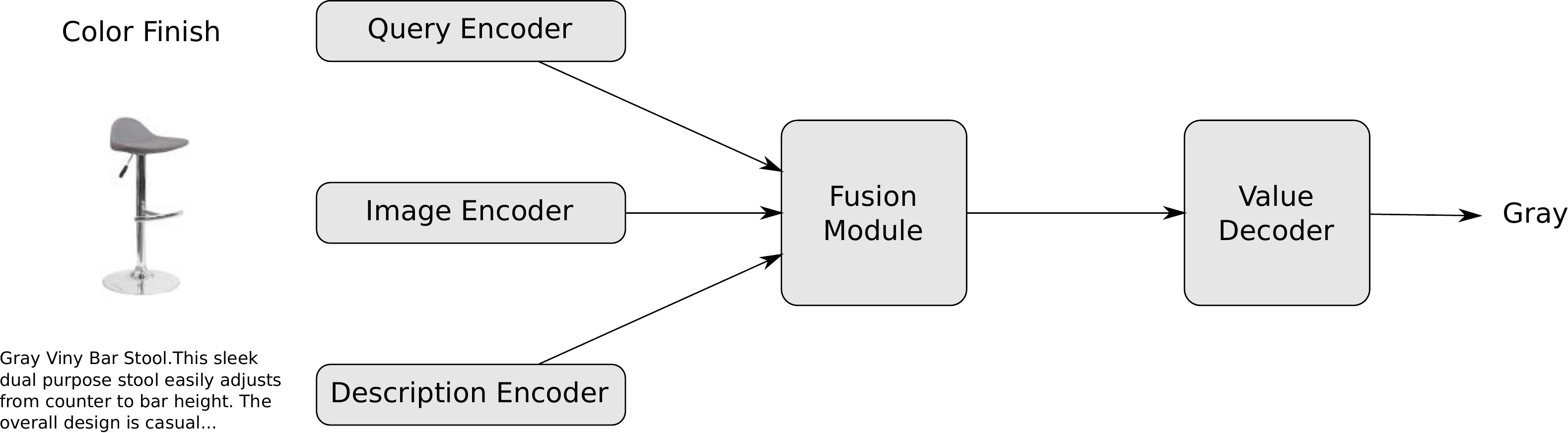}
    \caption{Basic architecture of the multimodal attribute extraction model.}
    \label{fig:model}
\end{figure}

\section{Multimodal Fusion Model}

In this section, we formulate a novel extraction model for the task that builds upon the architectures used recently in tasks such as image captioning, question answering, VQA, etc.
%embeds the query and input evidence into a real-valued vector space using
The model is composed of three separate modules: (1)~an encoding module that uses modern neural architectures to jointly embed the query, text, and images into a common latent space, (2)~a fusion module that combines these embedded vectors using an attribute-specific attention mechanism to a single dense vector, and (3)~a similarity-based value decoder which produces the final value prediction.
We provide an overview of this architecture in Figure~\ref{fig:model}.
% make a value prediction).
% %The embedding module .
% These embeddings are then combined in the fusion module  adapted from recently successful approaches for other multi-modal tasks such as image captioning and visual question answering.
% The mechanism assigns weights to each individual word and image, so that words/images that are more relevant to the given attribute query have higher weight.
% These weights are used to reduce the word and image embeddings to a single vector that is subsequently fed through 

%Note: In order to avoid superfluous notation in the following sections we will suppress the item index.

\paragraph{Encoding Module}
%\paragraph{Attribute and Value Embeddings}
We assign a dense embedding for each attribute and values, i.e. attribute $a$ is represented by a $k$-dimensional vector $\mathbf{c}_a$, and value $v$ by $\mathbf{c}_v$, where the vectors are learned during training. 
%
% Each attribute value pair $\langle a^j, v^j \rangle$
% the attribute and value are represented using a 1-of-N (one-hot) encodings $\mathbf{a}$ and $\mathbf{v}$ respectively.
% We randomly initialize weight matrices $\mathbf{W}_{a} \in \mathbb{R}^{|A| \times k}$ and $\mathbf{W}_{v} \in \mathbb{R}^{|V| \times k}$, which are used to embed the attribute and value vectors into a common vector space via matrix multiplication:
% \begin{equation}
% \begin{aligned}
%     \mathbf{c}_a &= \mathbf{a}\mathbf{W}_{a} \\
%     \mathbf{c}_v &= \mathbf{v} \mathbf{W}_{v}
% \end{aligned}
% \end{equation}
% The weights of $\mathbf{W}_{a}$ and $\mathbf{W}_{v}$ are subsequently learned during training.
%
%\paragraph{Text Embedding}
For textual description $D$, we first tokenize the text using the Stanford tokenizer~\citep{manning-EtAl:2014:P14-5}, followed by embedding all of the words using the Glove algorithm~\citep{pennington2014glove} on all of the descriptions in the training data.
We use the CNN architecture of \citet{kim2014convolutional}, that consists of CNN layers, max-pooling, and a fully-connected layer, to combine these pretrained embeddings to a single dense vector for the description, $\mathbf{c}_D$.
Embeddings of the images $I$ are also produced using convolutional neural networks.
Specifically, we obtain intermediate image representations using the output of the fc7 layer (after applying the ReLU non-linearity) of a pretrained 16-layer VGG model~\citep{simonyan2014very}.
We then feed the output through a fully connected layer to obtain a $k$-dimensional embedding for each image.
The final embedding $\mathbf{c}_I$ is produced by performing max-pooling over the image embeddings.

\paragraph{Fusion}
To fuse attribute embeddings $\mathbf{c}_a$ with the text and image embeddings, $\mathbf{c}_D$ and $\mathbf{c}_I$, we experiment with two different techniques.
The first, called \textbf{Concat}, is to concatenate the three of them and then feed them through a fully-connected layer, in order to produce the fused encoding $\mathbf{c}$.
The second approach, called \textbf{GMU} for gated multimodal unit~\citep{arevalo2017gated}, first fuses the attribute vector $\mathbf{c}_a$ with $\mathbf{c}_I$ and $\mathbf{c}_D$ independently using fully-connected layers, resulting in $\mathbf{c}^a_I$ and $\mathbf{c}^a_D$. We combine them by first creating gating vector $z = \sigma(\mathbf{W}_z [\mathbf{c}^a_D, \mathbf{c}^a_I])$, followed by, $\mathbf{c} = z \cdot \mathbf{c}^a_D + (1 - z) \cdot \mathbf{c}^a_I$.
For \emph{unimodal} baselines, the fusion module is replaced by a fully-connected layer.
% \begin{equation}
% \begin{aligned}
%     z &= \sigma(\mathbf{W}_z [\mathbf{c}_D, \mathbf{c}_I]) \\
%     \mathbf{c} &= z \cdot \mathbf{c}_D + (1 - z) \cdot \mathbf{c}_I
% \end{aligned}
% \end{equation}
%The case for image embeddings works similarly.

\paragraph{Loss Function}
We use a variant of the contrastive loss function introduced by \citet{chopra2005learning}. 
Let $\mathbf{c}$ denote the embedding produced by the fusion layer. Our goal is to produce an embedding which is close to the value embedding $\mathbf{c}^{+}_v$ (e.g. the one from the training example), and distant from other value embeddings $\mathbf{c}^{-}$. 
In order to measure closeness we use cosine similarity, denoted by $g$, %:
% \begin{equation}
%     g(\mathbf{c}, \mathbf{c}_v) = \frac{\mathbf{c}^T \mathbf{c}_v}{\left| \mathbf{c} \right| \left| \mathbf{c}_v \right |}
% \end{equation}
followed by a variant of squared hinge loss:
\begin{equation}
\begin{aligned}
    % \mathcal{L}^+(\mathbf{c}, \mathbf{c}^+_v) 
    %     &= (1 - g(\mathbf{c}, \mathbf{c}^+_v))^2 \\
    % \mathcal{L}^{-}(\mathbf{c}, \mathbf{c}^-_v)
    %     &=
    %         \begin{cases}
    %             g(\mathbf{c}, \mathbf{c}^-_v)^2 & \text{if} \, g(\mathbf{c}, \mathbf{c}^-_v) \geq 0 \\
    %             0 & \text{else}
    %         \end{cases} \\
    \mathcal{L}^+(\mathbf{c}, \mathbf{c}^+_v) 
        = (1 - g(\mathbf{c}, \mathbf{c}^+_v))^2; \hspace{5mm} &
    \mathcal{L}^{-}(\mathbf{c}, \mathbf{c}^-_v)
        =
            \begin{cases}
                g(\mathbf{c}, \mathbf{c}^-_v)^2 & \text{if} \, g(\mathbf{c}, \mathbf{c}^-_v) \geq 0 \\
                0 & \text{else}
            \end{cases} \\
    \mathcal{L}(\mathbf{c}, \mathbf{c}^+_v, \mathbf{c}^-_v)
        &= \mathcal{L}^+(\mathbf{c}, \mathbf{c}^+_v) + \mathcal{L}^-(\mathbf{c}, \mathbf{c}^-_v)
\end{aligned}
\end{equation}
where a negative value is sampled for each training example from the empirical distribution of value counts displayed in Figure~\ref{fig:hist}.
To obtain a value prediction given context, we identify the value with embedding $\mathbf{c}_v$ closest to the context embedding $\mathbf{c}$, according to cosine similarity $g$.

\section{Experiments}\label{section:experiments}

%\paragraph{Setup}
We evaluate on a subset of the MAE dataset consisting of the 100 most common attributes, %, and their corresponding values.
covering roughly 50\% of the examples in the overall MAE dataset.
To determine the relative effectiveness of the different modes of information, we train image and text only versions of the model described above.
Following the suggestions in~\citet{zhang2015sensitivity} we use a 600 unit single layer in our text convolutions, and a 5 word window size.
We apply dropout to the output of both the image and text CNNs before feeding the output through fully connected layers to obtain the image and text embeddings.
Employing a coarse grid search, we found models performed best using a large embedding dimension of $k=1024$.
Lastly, we explore multimodal models using both the \emph{Concat} and the \emph{GMU} strategies.
To evaluate models we use the hits@$k$ metric on the values. 
%Let $\hat{v}^j_{i,p}$ denote the top $p$ predictions for value $v_i^j$, then:
% $$ \text{Hits}@k = \frac{1}{N}\sum_{i \in \mathcal{I}}\mathds{1}(v_i^j \in \hat{v}^j_{i,p}) $$
% where $\mathds{1}$ denotes the indicator function. Note that $\text{Hits}@1$ is identical to accuracy.

%\paragraph{Results}
The results of our experiments are summarized in Table \ref{tbl:results}.
We include a simple most-common value model that always predicts the most-common value for a given attribute. %, as a baseline to compare other models against.
Observe that the performance of the image baseline model is almost identical to the most-common value model.
Similarly, the performance of the multimodal models is similar to the text baseline model.
Thus our models so far have been unable to effectively incorporate information from the image data.
These results show that the task is sufficiently challenging that even a complex neural model cannot solve the task, and thus is a ripe area for future research. 

Model predictions for the example shown in Figure~\ref{fig:example} are given in Table~\ref{fig:modeloutputs}, along with their similarity scores.
Observe that the predictions made by the current image baseline model are almost identical to the  most-common value model.
This suggests that our current image baseline model is essentially ignoring all of the image related information and instead learning to predict common values.

\begin{table}[t]
    \centering
    \caption{Baseline model results.}
    \label{tbl:results}
    \vskip -3mm
    \begin{tabular}{lcccc} \\\toprule
         & Hits@1 & Hits@5 & Hits@10 & Hits@20 \\
         \midrule
    Most-Common Value & 38.81 &	77.26 &	87.96 &	95.96 \\
    Image Baseline & 38.07 & 76.11 & 86.99 & 95.00 \\
    Text Baseline & 58.41 & 87.49 &	93.94 &	98.00 \\
    Multimodal Baseline - Concat & 59.48 & 87.33 & 93.23 & 97.07 \\
    Multimodal Baseline - GMU & 52.92 &	85.07 &	92.23 &	97.26 \\
    \bottomrule
    \end{tabular}
\end{table}

\begin{table}[t]
    \centering
    \caption{Top 5 predictions on the data in Figure~\ref{fig:example} when querying for \emph{color finish}.}
    \label{fig:modeloutputs}
    \small
    \begin{tabular}{lccccc}
    \toprule
        \bf Most-Common Value & \fcolorbox{Black}{White}{White} & \colorbox{Black}{\color{White}Black} & \colorbox{Gray}{\color{White}Stainless Steel} & \colorbox{LightGray}{Chrome} & \colorbox{DarkGray}{\color{White}Gray}\\ \addlinespace
        \multirow{ 2}{*}{\bf Text Baseline} &\colorbox{DarkGray}{\color{White}Gray} & \colorbox{LightGray}{Silver} & \colorbox{DarkGray}{\color{White}Grey} & \fcolorbox{Black}{White}{White} & \colorbox{Beige}{Beige} \\
            &0.84 & 0.63 & 0.60 & 0.60 & 0.58\\ \addlinespace
        \multirow{ 2}{*}{\bf Image Baseline} & \fcolorbox{Black}{White}{White} & \colorbox{Black}{\color{White}Black} &

 \colorbox{DarkBlue}{\color{White}Blue} &\colorbox{DarkGray}{\color{White}Gray} & \colorbox{Brown}{\color{White}Brown} \\
            &0.81 & 0.70 & 0.63 & 0.62 & 0.59 \\ \addlinespace
        \multirow{ 2}{*}{\bf Multimodal Baseline - Concat}  & \colorbox{DarkGray}{\color{White}Gray}  & \colorbox{BrickRed}{\color{White}Red} & \colorbox{LightGreen}{Green} & \colorbox{DarkGray}{\color{White}Grey} & \colorbox{DarkBlue}{\color{White}Blue} \\
            & 0.84 & 0.71 & 0.71 & 0.71 & 0.70 \\ \addlinespace
        \multirow{ 2}{*}{\bf Multimodal Baseline - GMU} & \colorbox{DarkGray}{\color{White}Gray} & \colorbox{DarkBlue}{\color{White}Blue} & \colorbox{Brown}{\color{White}Brown} & \colorbox{LightGreen}{Green} & \colorbox{BrickRed}{\color{White}Red} \\
            &0.85 & 0.71 & 0.69 & 0.68 & 0.67 \\
    \bottomrule
    \end{tabular}
    \vskip -5mm
\end{table}

\section{Related Work}

Our work is related to, and builds upon, a number of existing approaches.

%\paragraph{Datasets}
The introduction of large curated datasets has driven progress in many fields of machine learning. % over the years. 
Notable examples include:
The Penn Treebank~\citep{marcus1993building} for syntactic parsing models,
Imagenet~\citep{deng2009imagenet} for object recognition,
Flickr30k~\citep{young2014image} and MS COCO~\citep{lin2014microsoft} for image captioning,
SQuAD~\citep{rajpurkar2016squad} for question answering
and VQA~\citep{VQA} for visual question answering.
Despite the interest in related tasks, there is currently no publicly available dataset for attribute extraction, let alone multimodal attribute extraction.
This creates a high barrier to entry as anyone interested in attribute extraction must go through the expensive and time-consuming process of acquiring a dataset.
Furthermore, there is no way to compare the effectiveness of different techniques.
%We release the \ds (\dsabbrev) 
Our dataset aims to address this concern. %, detailed in section~\ref{section:mae}.

%\paragraph{Multimodal Machine Learning}

Recently, there has been renewed interest in multimodal machine learning problems. % in the past few years.
% Below - popular multimodal models, but largely irrelevant
%\citet{vinyals2015show} demonstrate an effective image captioning system, which uses a convolutional neural network to encode an image into a fixed vector which is used as the first input into an long short term memory (LSTM)~\citep{hochreiter1997long} decoder which produces sequence of words forming the output caption.
\citet{vinyals2015show} demonstrate an effective image captioning system that uses a CNN to encode an image which is used as the input to an LSTM~\citep{hochreiter1997long} decoder, producing the output caption.
This encoder-decoder architecture forms the basis for successful approaches to other multimodal problems such as visual question answering~\cite{anderson2017bottom}.
% These works demonstrate how data of one modality can be translated to another.
% Below - less popular multimodal models but more related to our work on the basis of fusion
Another body of work focuses on the problem of unifying information from different modes of information.
%In multimodal distributional semantics, the goal is to produce semantic vectors using a combination of images and text.
\citet{kiela2014learning} propose to concatenate together the output of a text-based distributional model (such as word2vec~\cite{mikolov2013distributed}) with an encoding produced from a CNN applied to images of the word.
\citet{lazaridou2015combining} demonstrate an alternative approach to concatenation, where instead the a word embedding is learned that minimizes a joint loss function involving context-prediction and image reconstruction losses.
Another alternative to concatenation is the gated multimodal unit (GMU) proposed in \cite{arevalo2017gated}.
We investigate the performance of different techniques for combining image and text data for product attribute extraction in section \ref{section:experiments}.

%\paragraph{Product Attribute Extraction}

To our knowledge, we are the first to study the problem of attribute extraction from multimodal data.
However the  problem of attribute extraction from text is well studied.
\citet{Ghani:2006} treat attribute extraction of retail products as a form of named entity recognition. %represent  by a collection of attribute-value pairs, and 
They predefine a list of attributes to extract and train a Na\"ive Bayes model on a manually labeled seed dataset to extract the corresponding values.
% They also demonstrate that additional performance can be gained by augmenting the seed dataset with unlabeled data and and using Expectation-Maximization (EM).
\citet{putthividhya2011bootstrapped} build on this work by bootstrapping to expand the seed list, and evaluate more complicated models such as HMMs, MaxEnt, SVMs, and CRFs. %instead using a , technique , the performance of 
To mitigate the introduction noisy labels when using semi-supervised techniques, \citet{more2016attribute} incorporates crowdsourcing to manually accept or reject the newly introduced labels. % are either accepted or rejected by human workers using crowd sourcing.
% where 
%
One major drawback of these approaches is that they require manually labelled seed data to construct the knowledge base of attribute-value pairs, which can be quite expensive for a large number of attributes. %if one wants to train a system that works 
\citet{bing2012unsupervised} address this problem by using an unsupervised, LDA-based approach to generate word classes from reviews, followed by aligning them to the product description. %an unsupervised, LDA-based method for extracting product attribute-value pairs.
%The method uses Latent Dirichlet Allocation (LDA)~\citep{blei2003latent} to generate word classes from product review data.
%A training dataset is constructed by matching words in classes to those in product descriptions.
\citet{shinzato2013unsupervised} propose to extract attribute-value pairs from structured data on product pages, such as HTML tables, and lists, to construct the KB.
This is essentially the approach used to construct the knowledge base of attribute-value pairs used in our work, which is automatically performed by Diffbot's Product API.

\section{Conclusions and Future Work}
% Following  SQuAD for inspiration
In order to kick start research on multimodal information extraction problems, we introduce the multimodal attribute extraction dataset, an attribute extraction dataset derived from a large number of e-commerce websites.
\dsabbrev features images, textual descriptions, and attribute-value pairs for a diverse set of products.
Preliminary data from an Amazon Mechanical Turk study demonstrates that both modes of information are beneficial to attribute extraction. 
We measure the performance of a collection of baseline models, and observe that reasonably high accuracy can be obtained using only text. 
However, we are unable to train off-the-shelf methods to effectively leverage image data.

There are a number of exciting avenues for future research.
We are interested in performing a more comprehensive crowdsourcing study to identify the ways in which different evidence forms are useful, and in order to create \emph{clean} evaluation data.
As this dataset brings up interesting challenges in multimodal machine learning, we will explore a variety of novel architectures that are able to combine the different forms of evidence effectively to accurately extract the attribute values.
Finally, we are also interested in exploring other aspects of knowledge base construction that may benefit from multimodal reasoning, such as relational prediction, entity linking, and disambiguation.

\subsubsection*{Acknowledgments}

The authors are grateful to Diffbot for generously providing API access for the \dsabbrev dataset, as well as support for this research.

\bibliography{main}

\begin{thebibliography}{25}
\providecommand{\natexlab}[1]{#1}
\providecommand{\url}[1]{\texttt{#1}}
\expandafter\ifx\csname urlstyle\endcsname\relax
  \providecommand{\doi}[1]{doi: #1}\else
  \providecommand{\doi}{doi: \begingroup \urlstyle{rm}\Url}\fi

\bibitem[Anderson et~al.(2017)Anderson, He, Buehler, Teney, Johnson, Gould, and
  Zhang]{anderson2017bottom}
P.~Anderson, X.~He, C.~Buehler, D.~Teney, M.~Johnson, S.~Gould, and L.~Zhang.
\newblock Bottom-up and top-down attention for image captioning and vqa.
\newblock \emph{arXiv preprint arXiv:1707.07998}, 2017.

\bibitem[Antol et~al.(2015)Antol, Agrawal, Lu, Mitchell, Batra, Zitnick, and
  Parikh]{VQA}
S.~Antol, A.~Agrawal, J.~Lu, M.~Mitchell, D.~Batra, C.~L. Zitnick, and
  D.~Parikh.
\newblock {VQA}: {V}isual {Q}uestion {A}nswering.
\newblock In \emph{International Conference on Computer Vision (ICCV)}, 2015.

\bibitem[Arevalo et~al.(2017)Arevalo, Solorio, Montes-y G{\'o}mez, and
  Gonz{\'a}lez]{arevalo2017gated}
J.~Arevalo, T.~Solorio, M.~Montes-y G{\'o}mez, and F.~A. Gonz{\'a}lez.
\newblock Gated multimodal units for information fusion.
\newblock \emph{arXiv preprint arXiv:1702.01992}, 2017.

\bibitem[Bing et~al.(2012)Bing, Wong, and Lam]{bing2012unsupervised}
L.~Bing, T.-L. Wong, and W.~Lam.
\newblock Unsupervised extraction of popular product attributes from web sites.
\newblock In \emph{Asia Information Retrieval Symposium}, pages 437--446.
  Springer, 2012.

\bibitem[Chopra et~al.(2005)Chopra, Hadsell, and LeCun]{chopra2005learning}
S.~Chopra, R.~Hadsell, and Y.~LeCun.
\newblock Learning a similarity metric discriminatively, with application to
  face verification.
\newblock In \emph{Computer Vision and Pattern Recognition, 2005. CVPR 2005.
  IEEE Computer Society Conference on}, volume~1, pages 539--546. IEEE, 2005.

\bibitem[Deng et~al.(2009)Deng, Dong, Socher, Li, Li, and
  Fei-Fei]{deng2009imagenet}
J.~Deng, W.~Dong, R.~Socher, L.-J. Li, K.~Li, and L.~Fei-Fei.
\newblock Imagenet: A large-scale hierarchical image database.
\newblock In \emph{Computer Vision and Pattern Recognition, 2009. CVPR 2009.
  IEEE Conference on}, pages 248--255. IEEE, 2009.

\bibitem[Ghani et~al.(2006)Ghani, Probst, Liu, Krema, and Fano]{Ghani:2006}
R.~Ghani, K.~Probst, Y.~Liu, M.~Krema, and A.~Fano.
\newblock Text mining for product attribute extraction.
\newblock \emph{SIGKDD Explor. Newsl.}, 8\penalty0 (1):\penalty0 41--48, June
  2006.
\newblock ISSN 1931-0145.
\newblock \doi{10.1145/1147234.1147241}.
\newblock URL \url{http://doi.acm.org/10.1145/1147234.1147241}.

\bibitem[Hochreiter and Schmidhuber(1997)]{hochreiter1997long}
S.~Hochreiter and J.~Schmidhuber.
\newblock Long short-term memory.
\newblock \emph{Neural computation}, 9\penalty0 (8):\penalty0 1735--1780, 1997.

\bibitem[Kiela and Bottou(2014)]{kiela2014learning}
D.~Kiela and L.~Bottou.
\newblock Learning image embeddings using convolutional neural networks for
  improved multi-modal semantics.
\newblock In \emph{Empirical Methods for Natural Language Processing (EMNLP)},
  2014.

\bibitem[Kim(2014)]{kim2014convolutional}
Y.~Kim.
\newblock Convolutional neural networks for sentence classification.
\newblock \emph{arXiv preprint arXiv:1408.5882}, 2014.

\bibitem[Lazaridou et~al.(2015)Lazaridou, Pham, and
  Baroni]{lazaridou2015combining}
A.~Lazaridou, N.~T. Pham, and M.~Baroni.
\newblock Combining language and vision with a multimodal skip-gram model.
\newblock \emph{arXiv preprint arXiv:1501.02598}, 2015.

\bibitem[Lin et~al.(2014)Lin, Maire, Belongie, Hays, Perona, Ramanan,
  Doll{\'a}r, and Zitnick]{lin2014microsoft}
T.-Y. Lin, M.~Maire, S.~Belongie, J.~Hays, P.~Perona, D.~Ramanan,
  P.~Doll{\'a}r, and C.~L. Zitnick.
\newblock Microsoft coco: Common objects in context.
\newblock In \emph{European conference on computer vision}, pages 740--755.
  Springer, 2014.

\bibitem[Manning et~al.(2014)Manning, Surdeanu, Bauer, Finkel, Bethard, and
  McClosky]{manning-EtAl:2014:P14-5}
C.~D. Manning, M.~Surdeanu, J.~Bauer, J.~Finkel, S.~J. Bethard, and
  D.~McClosky.
\newblock The {Stanford} {CoreNLP} natural language processing toolkit.
\newblock In \emph{Association for Computational Linguistics (ACL) System
  Demonstrations}, pages 55--60, 2014.
\newblock URL \url{http://www.aclweb.org/anthology/P/P14/P14-5010}.

\bibitem[Marcus et~al.(1993)Marcus, Marcinkiewicz, and
  Santorini]{marcus1993building}
M.~P. Marcus, M.~A. Marcinkiewicz, and B.~Santorini.
\newblock Building a large annotated corpus of english: The penn treebank.
\newblock \emph{Computational linguistics}, 19\penalty0 (2):\penalty0 313--330,
  1993.

\bibitem[Mikolov et~al.(2013)Mikolov, Sutskever, Chen, Corrado, and
  Dean]{mikolov2013distributed}
T.~Mikolov, I.~Sutskever, K.~Chen, G.~S. Corrado, and J.~Dean.
\newblock Distributed representations of words and phrases and their
  compositionality.
\newblock In \emph{Advances in neural information processing systems}, pages
  3111--3119, 2013.

\bibitem[More(2016)]{more2016attribute}
A.~More.
\newblock Attribute extraction from product titles in ecommerce.
\newblock \emph{arXiv preprint arXiv:1608.04670}, 2016.

\bibitem[Pennington et~al.(2014)Pennington, Socher, and
  Manning]{pennington2014glove}
J.~Pennington, R.~Socher, and C.~D. Manning.
\newblock Glove: Global vectors for word representation.
\newblock In \emph{Empirical Methods in Natural Language Processing (EMNLP)},
  pages 1532--1543, 2014.
\newblock URL \url{http://www.aclweb.org/anthology/D14-1162}.

\bibitem[Putthividhya and Hu(2011)]{putthividhya2011bootstrapped}
D.~P. Putthividhya and J.~Hu.
\newblock Bootstrapped named entity recognition for product attribute
  extraction.
\newblock In \emph{Proceedings of the Conference on Empirical Methods in
  Natural Language Processing}, pages 1557--1567. Association for Computational
  Linguistics, 2011.

\bibitem[Rajpurkar et~al.(2016)Rajpurkar, Zhang, Lopyrev, and
  Liang]{rajpurkar2016squad}
P.~Rajpurkar, J.~Zhang, K.~Lopyrev, and P.~Liang.
\newblock Squad: 100,000+ questions for machine comprehension of text.
\newblock \emph{arXiv preprint arXiv:1606.05250}, 2016.

\bibitem[Shinzato and Sekine(2013)]{shinzato2013unsupervised}
K.~Shinzato and S.~Sekine.
\newblock Unsupervised extraction of attributes and their values from product
  description.
\newblock In \emph{International Joint Conference on Natural Language
  Processing (IJCNLP)}, 2013.

\bibitem[Simonyan and Zisserman(2014)]{simonyan2014very}
K.~Simonyan and A.~Zisserman.
\newblock Very deep convolutional networks for large-scale image recognition.
\newblock \emph{arXiv preprint arXiv:1409.1556}, 2014.

\bibitem[Szegedy et~al.(2016)Szegedy, Vanhoucke, Ioffe, Shlens, and
  Wojna]{szegedy2016rethinking}
C.~Szegedy, V.~Vanhoucke, S.~Ioffe, J.~Shlens, and Z.~Wojna.
\newblock Rethinking the inception architecture for computer vision.
\newblock In \emph{Proceedings of the IEEE Conference on Computer Vision and
  Pattern Recognition}, pages 2818--2826, 2016.

\bibitem[Vinyals et~al.(2015)Vinyals, Toshev, Bengio, and
  Erhan]{vinyals2015show}
O.~Vinyals, A.~Toshev, S.~Bengio, and D.~Erhan.
\newblock Show and tell: A neural image caption generator.
\newblock In \emph{Proceedings of the IEEE conference on computer vision and
  pattern recognition}, pages 3156--3164, 2015.

\bibitem[Young et~al.(2014)Young, Lai, Hodosh, and Hockenmaier]{young2014image}
P.~Young, A.~Lai, M.~Hodosh, and J.~Hockenmaier.
\newblock From image descriptions to visual denotations: New similarity metrics
  for semantic inference over event descriptions.
\newblock \emph{Transactions of the Association for Computational Linguistics},
  2:\penalty0 67--78, 2014.

\bibitem[Zhang and Wallace(2015)]{zhang2015sensitivity}
Y.~Zhang and B.~Wallace.
\newblock A sensitivity analysis of (and practitioners' guide to) convolutional
  neural networks for sentence classification.
\newblock \emph{arXiv preprint arXiv:1510.03820}, 2015.

\end{thebibliography}

\end{document}